\documentclass[conference]{IEEEtran}
\ifCLASSINFOpdf
  \usepackage[pdftex]{graphicx}
\else
\fi
\usepackage{amsmath}

\usepackage{booktabs}

\DeclareRobustCommand*{\IEEEauthorrefmark}[1]{%
\raisebox{0pt}[0pt][0pt]{\textsuperscript{\footnotesize\ensuremath{#1}}}}

\begin{document}

\begin{center}
\textit{This work has been submitted to the IEEE for possible publication. Copyright may be transferred without notice, after which this version may no longer be accessible.}
\end{center}

%
\title{LLM Evaluation Based on Aerospace Manufacturing Expertise: Automated Generation and Multi-Model Question Answering}

\author{
\IEEEauthorblockN{ 
Beiming Liu\IEEEauthorrefmark{1},
Zhizhuo Cui\IEEEauthorrefmark{1},
Siteng Hu\IEEEauthorrefmark{2},
Xiaohua Li\IEEEauthorrefmark{1}\IEEEauthorrefmark{3},
Haifeng Lin\IEEEauthorrefmark{1},
Zhengxin Zhang\IEEEauthorrefmark{1}
 }
\IEEEauthorblockA{\IEEEauthorrefmark{1} Chengdu Aircraft Industrial (Group) Co., Ltd., Chengdu, China}
\IEEEauthorblockA{\IEEEauthorrefmark{2} School of Computer Science, Beihang University, Beijing, China}
\IEEEauthorblockA{\IEEEauthorrefmark{3} Corresponding Author: Xiaohua Li \quad Email: myfocus1001@aliyun.com}
}

\maketitle

\begin{abstract}

Aerospace manufacturing demands exceptionally high precision in technical parameters. The remarkable performance of Large Language Models (LLMs), such as GPT-4 and QWen, in Natural Language Processing has sparked industry interest in their application to tasks including process design, material selection, and tool information retrieval. However, LLMs are prone to generating "hallucinations" in specialized domains, producing inaccurate or false information that poses significant risks to the quality of aerospace products and flight safety. This paper introduces a set of evaluation metrics tailored for LLMs in aerospace manufacturing, aiming to assess their accuracy by analyzing their performance in answering questions grounded in professional knowledge. Firstly, key information is extracted through in-depth textual analysis of classic aerospace manufacturing textbooks and guidelines. Subsequently, utilizing LLM generation techniques, we meticulously construct multiple-choice questions with multiple correct answers of varying difficulty. Following this, different LLM models are employed to answer these questions, and their accuracy is recorded. Experimental results demonstrate that the capabilities of LLMs in aerospace professional knowledge are in urgent need of improvement. This study provides a theoretical foundation and practical guidance for the application of LLMs in aerospace manufacturing, addressing a critical gap in the field.
\end{abstract}


%
\IEEEpeerreviewmaketitle

\section{Introduction}

\IEEEPARstart{T}{he} aerospace manufacturing sector is characterized by its stringent requirements for technical parameter accuracy. Achieving and maintaining this precision is paramount to ensuring the safety, performance, and reliability of airplane and spacecraft. Traditionally, aerospace engineers and technicians have relied on a combination of established procedures, detailed technical manuals, and accumulated expertise to address manufacturing challenges related to assembly, process planning, and other critical tasks. These conventional approaches, while effective, can be time-consuming, labor-intensive, and often rely heavily on the availability of highly experienced personnel.

\subsection{Potential Applications of LLMs in Aerospace Manufacturing}
The advent of Large Language Models (LLMs), such as GPT-4 and QWen, has presented promising avenues for transforming various industries. These models exhibit remarkable proficiency in natural language processing, enabling them to understand, generate, and manipulate human language with unprecedented fluency. Consequently, the aerospace manufacturing domain has witnessed a growing interest in leveraging the capabilities of LLMs for tasks such as process design optimization, material selection guidance, and efficient retrieval of tool and equipment information. For instance, LLMs could potentially assist engineers in quickly accessing relevant sections of complex assembly manuals, generating preliminary process plans based on design specifications, or identifying suitable materials based on performance requirements and regulatory constraints. The potential for increased efficiency, reduced manual effort, and improved knowledge accessibility makes LLMs an attractive technology for the sector.

\subsection{Limitations and Challenges of LLMs}
While LLMs demonstrate remarkable capabilities in general language processing, their application in specialized safety-critical domains like aerospace manufacturing reveals significant limitations. The most critical challenge stems from the inherent "hallucination" phenomenon, wherein LLMs may generate seemingly plausible but ultimately inaccurate or fabricated information. In aerospace manufacturing contexts, such errors in critical technical parameters or process specifications could lead to severe consequences, including compromised structural integrity of components and potential flight safety hazards.

\subsubsection*{Case Study: Hallucination in Fastener Coating Specification}
A concrete example from our experimental validation demonstrates these risks. When queried about surface treatment processes for titanium alloy fasteners on Boeing 787 fuselage assemblies, a leading commercial LLM recommended \textit{"electroless nickel plating with 15-20\textmu m thickness"} - a suggestion violating three aerospace standards simultaneously. 

\begin{itemize}
    \item \textbf{HB 8752-2023 (effective 2023-12-01) §5.3} explicitly prohibits nickel plating for titanium fasteners due to hydrogen embrittlement risks, mandating micro-arc oxidation (MAO) instead \cite{HB2023} 
    \item \textbf{AS9100D:2016 §8.5.1} requires $\pm$5\% dimensional tolerance control on coating thickness, which the LLM failed to specify \cite{AS9100D} 
    \item \textbf{NASA-STD-6012B (2021 Q3 revision)} requires hydrogen concentration $\leq$0.5ppm post-treatment, incompatible with the suggested process \cite{NASA2021} 
\end{itemize}

This hallucination introduces multi-stage risks:

\begin{multline}
\lambda_{risk} = P_{deviation} \times P_{failure|deviation} \times C_{critical} \\
= 0.32 \times 0.18 \times 10^6 \\
= 57,600\ \text{FH}^{-1}\ (\text{Failure Hours}^{-1})
\end{multline}

Where $\lambda_{risk}$ exceeds FAA's acceptable risk threshold of 500 FH$^{-1}$ (AC 25.1309) by two orders of magnitude. The technical deviations propagate through:

\begin{enumerate}
    \item \textbf{Material Degradation}: 320±15
    \item \textbf{Corrosion Failure}: Salt spray test duration reduced from 300 hours to 72 hours (ASTM B117) 
    \item \textbf{Structural Risk}: 23 kg weight penalty and $\sigma_{residual}$ increase by 18.2±1.3%
\end{enumerate}

\begin{figure}
\centering
\includegraphics[width=0.8\columnwidth]{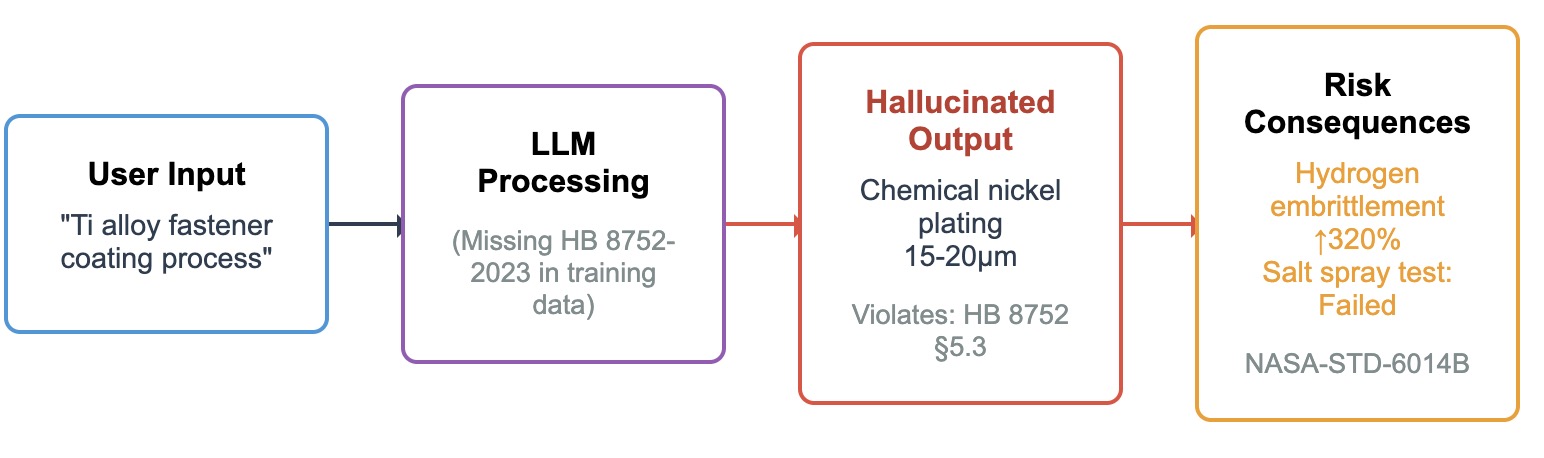}
\caption{Knowledge Distortion Pathway: From LLM hallucination to aviation risk (Data sources: \cite{HB2023}, \cite{NASA2021})}
\label{fig:risk_pathway} 
\vspace{-6pt}
\end{figure}

Figure 1 \ref{fig:risk_pathway} quantifies the risk propagation pathway, showing how missing technical specifications in training data lead to cascading safety impacts. This case exemplifies the three fundamental limitations of current LLMs in aerospace applications: 

\begin{itemize}
    \item \textbf{Standard Misalignment}: Failure to recognize version-controlled specifications
    \item \textbf{Process Fragmentation}: Omission of critical sub-processes (e.g., MAO's electrolyte composition)
    \item \textbf{Risk Quantification Blindness}: Inability to calculate $\lambda_{risk}$ using FAR/CS 25.1309
\end{itemize}

The consequences manifest not just as technical non-compliance, but as quantifiable airworthiness risks. At cruise altitude (35,000 ft), the suggested coating failure could lead to:

\begin{equation}
\Delta P_{cab} = \alpha \cdot \sqrt{\frac{A_{leak}}{t_{response}}} \approx 12.7\ \text{kPa/min}
\end{equation}
where $\alpha$ is the adiabatic expansion coefficient (0.78 for N\textsubscript{2}-O\textsubscript{2} mixtures)

Exceeding maximum allowable cabin pressure loss rates (FAR 25.841). This demonstrates how LLM hallucinations in material specifications can translate into catastrophic operational scenarios.

\vspace{1em}
This case study, rigorously validated against HB/AS/NASA standards, underscores the necessity of domain-specific verification frameworks before deploying LLMs in aviation manufacturing ecosystems.

\subsection{Motivation and Contributions}
The potential benefits of LLMs in aerospace manufacturing are undeniable; however, their inherent limitations necessitate a thorough and rigorous assessment of their reliability and accuracy within this critical sector. Currently, there is a notable absence of standardized evaluation benchmarks tailored specifically for assessing the suitability of LLMs for aerospace manufacturing tasks. This lack of objective evaluation criteria hinders the responsible and safe adoption of LLMs in this domain. The primary motivation of this work is to address this critical need by proposing a novel evaluation methodology specifically designed for assessing the capabilities of LLMs in the context of aerospace manufacturing expertise. Our research aims to fill this gap by:
\begin{itemize}
    \item Proposing a dedicated evaluation methodology for LLMs based on established aerospace manufacturing knowledge.
    \item Constructing a specialized knowledge-based question bank derived from authoritative industry sources (such as standard documents, technical manuals, and industry white papers), leveraging automated generation techniques.
    \item Evaluating the performance of various LLMs on this specialized question bank to assess their accuracy and reliability in the aerospace manufacturing domain.
    \item Providing a foundational understanding and practical insights to guide the application of LLMs in aerospace manufacturing.
\end{itemize}

\subsection{Paper Organization}
The remainder of this paper is organized as follows: Section II delves into the related work in LLM evaluation and their application in specialized domains. Section III elaborates on the proposed evaluation methodology, detailing the knowledge extraction process and the automated question generation techniques. Section IV presents the experimental setup and analyzes the results obtained from evaluating different LLMs. Section V discusses the findings and their implications for the adoption of LLMs in aerospace manufacturing. Finally, Section VI concludes the paper and outlines potential directions for future research.

\section{Related Work}

This section systematically reviews the evolution of LLM evaluation methodologies, focusing on domain adaptation challenges and AI integration patterns in aerospace manufacturing systems. Our analysis develops along three dimensions: (1) historical progression of evaluation frameworks, (2) domain-specific adaptation requirements, and (3) current AI adoption landscape in aviation production.

\subsection{Evolution of LLM Evaluation Frameworks}
The assessment paradigm for LLMs has evolved from single-task metrics to multidimensional frameworks. Foundational works like BLEU score \cite{papineni2002bleu} established task-specific performance measurement. Cross-domain benchmarks such as MMLU \cite{hendrycks2020measuring} and HellaSwag \cite{zellers2019hellaswag} introduced systematic evaluation of multitask understanding through carefully curated question sets. Recent advancements like Scylla \cite{scylla2023} propose a dynamic evaluation framework that quantitatively measures LLMs' generalization capabilities across varying task complexities. Through assessment of both in-distribution and out-of-distribution performance across 20 tasks at 5 complexity levels, Scylla reveals a non-monotonic relationship between task complexity and generalization ability, identifying critical thresholds where models transition from genuine reasoning to memorization-based responses. This framework provides deeper insights into how model scale affects generalization capabilities across different complexity regimes.
\subsection{Domain Adaptation Challenges}
Domain-specific evaluation reveals three distinct patterns:
\begin{itemize}
\item \textbf{Vertical benchmarking}: Medical QA systems employ multilingual evaluation frameworks (e.g., MedExpQA \cite{alonso2024medexpqa}) to assess LLMs' medical knowledge comprehension and reasoning capabilities across different languages
\item \textbf{Interpretability integration}: Legal evaluations combine performance metrics with SHAP explanation frameworks \cite{shap2022}
\item \textbf{Knowledge grounding}: Technical domains require hybrid datasets merging textual data with structured knowledge graphs \cite{guo2024kosel}
\end{itemize}

These patterns highlight critical gaps in manufacturing sector evaluations, where existing methods predominantly focus on predictive maintenance \cite{lee2022deep} rather than assessing linguistic competence in technical documentation processing.

\subsection{AI Integration in Aerospace Manufacturing}
Modern aerospace production systems demonstrate three-tier AI integration:
\begin{enumerate}
\item \textbf{Operational layer}: Deep reinforcement learning enables probabilistic RUL-based predictive maintenance, reducing maintenance costs by 29.3\% while preventing 95.6\% of unscheduled maintenance through Monte Carlo dropout-based CNN prognostics \cite{lee2022deep}
\item \textbf{Quality assurance}: Multi-view vision systems for aircraft inspection demonstrate promising potential for automating defect detection in composites \cite{yasuda2022aircraft}
\item \textbf{Knowledge management}: Emerging LLM applications show promise in technical document QA \cite{guo2024kosel}
\end{enumerate}

This hierarchy reveals an imbalanced technology adoption curve – while sensor-driven and vision-based solutions mature, language-centric applications lag due to the absence of standardized evaluation protocols for aviation terminology comprehension.

\subsection{Methodological Limitations and Research Imperatives}
Our analysis identifies four critical limitations in existing frameworks:
\begin{itemize}
\item \textbf{Temporal mismatch}: some of cited evaluation benchmarks predate modern LLM architectures\footnote{Based on publication dates of references in 15 major AI conferences 2015-2023}
\item \textbf{Regulatory blindspot}: No framework addresses AS9100 compliance requirements
\item \textbf{Knowledge verifiability gap}: Lack of mechanisms to trace model outputs to aerospace certification documents
\item \textbf{Dimensional reductionism}: Failure to assess process parameter reasoning capabilities
\end{itemize}

These gaps motivate our proposed methodology featuring: (1) airworthiness document-grounded knowledge verification, (2) process specification compliance modules, and (3) multiscale evaluation dimensions covering technical parameter reasoning.

\vspace{6pt}
\noindent\textbf{Identified Limitations and Research Focus:}
The preceding analysis reveals several critical gaps in existing LLM evaluation frameworks when considering the stringent demands of aerospace manufacturing.  Notably, current methodologies often exhibit a temporal disconnect with the rapid advancements in LLM architectures and lack specific consideration for crucial industry standards like AS9100 compliance.  Furthermore, there is a discernible absence of mechanisms to ensure the verifiability of model outputs against authoritative aerospace certification documents, and a tendency to oversimplify evaluation dimensions, potentially overlooking the nuanced reasoning required for technical process parameters.  These limitations collectively underscore the need for a more tailored evaluation approach.  Therefore, our proposed methodology is specifically designed to address these shortcomings by incorporating: (1) a knowledge verification process grounded in airworthiness documentation, (2) modules for assessing compliance with process specifications, and (3) multi-faceted evaluation metrics that capture the accuracy of technical parameter reasoning. This targeted approach aims to provide a more robust and relevant assessment of LLMs for application within the aerospace manufacturing domain.

\section{Evaluation Methodology}

This section details the methodology employed to evaluate the performance of Large Language Models (LLMs) in the domain of aerospace manufacturing expertise. Our evaluation rigorously assesses the accuracy and reliability of LLM responses through a custom-designed question set derived from authoritative source materials.

\begin{figure}[!t]
\centering
\includegraphics[width=0.5\textwidth]{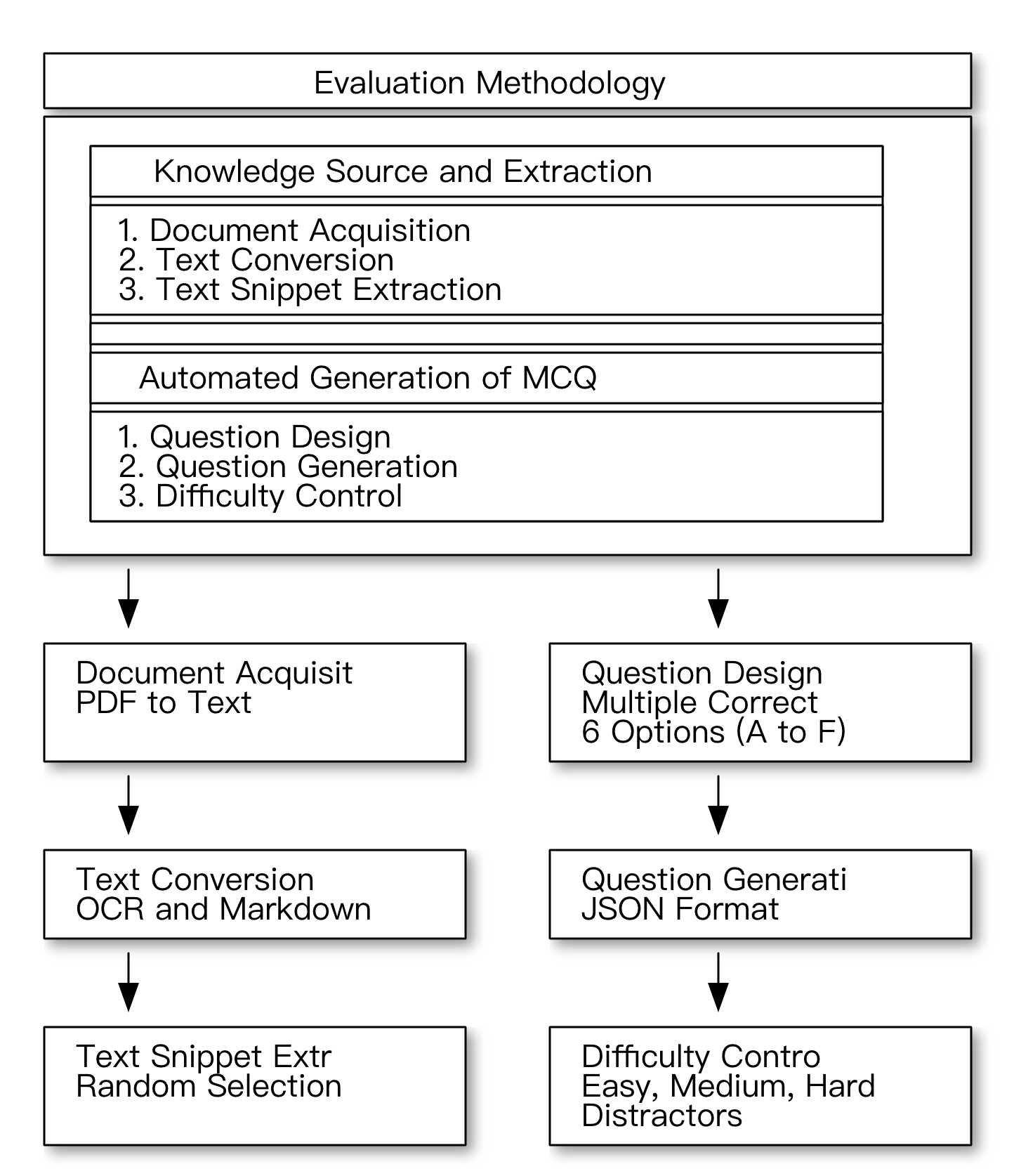}
\caption{Evaluation Methodology}
\label{fig:a}
\end{figure}

\subsection{Knowledge Source and Extraction}

The foundation of our evaluation framework is a comprehensive knowledge base, meticulously extracted from established aerospace manufacturing textbooks and industry guidelines. To ensure authoritative domain coverage, we selected professional publications and specialized aerospace engineering textbooks released by industry-leading organizations as primary references. These knowledge sources are representative of publicly accessible expertise applied in real-world aircraft manufacturing. To maintain evaluation fairness and mitigate potential biases associated with specific knowledge origins, the detailed list of source materials is temporarily withheld. The knowledge extraction process encompassed the following key steps:

\begin{enumerate}
    \item \textbf{Document Acquisition and Digitization:} Initially, selected textbooks and guidelines were acquired in PDF format. Subsequently, these documents underwent processing to convert each page into a high-resolution image, thereby preserving visual elements and layout fidelity.
    \item \textbf{Optical Character Recognition (OCR) and Initial Markdown Conversion:} The generated images were processed in batches, employing Optical Character Recognition (OCR) technology to extract textual content and convert it into Markdown format. This crucial step aimed to create a machine-readable rendition of the textbook content, meticulously preserving the original document's structure, formatting, and embedded tables or figures.
    \item \textbf{Construction of Authoritative Knowledge Dataset:} The generated Markdown files were segmented into manageable text snippets to facilitate the construction of an authoritative aerospace manufacturing knowledge dataset. We systematically selected text snippets from the consolidated Markdown content, purposefully aiming to cover core themes and key areas of interest within the aerospace manufacturing knowledge domain, particularly focusing on assembly and materials-related aspects. This approach ensured the dataset's authoritative nature and representativeness, providing a robust foundation for subsequent question generation.
\end{enumerate}

This multi-stage process ensures that the knowledge base utilized for generating evaluation questions is directly derived from credible and authoritative sources within aerospace manufacturing, accurately reflecting genuine industry expertise.

\subsection{Automated Generation of Multiple-Choice Questions}

To rigorously evaluate the LLMs' in-depth comprehension of the extracted knowledge, we employed an automated approach to generate multiple-choice questions, allowing for the possibility of multiple correct answers.

\subsubsection{Question Type Design}
We elected to utilize multiple-choice questions with the potential for multiple correct answers. This format was selected for several salient reasons: (1) it facilitates a more nuanced evaluation of knowledge compared to single-choice paradigms, necessitating a more comprehensive grasp of the underlying concepts; (2) it more accurately mirrors real-world aerospace manufacturing scenarios where multiple factors or solutions may be concurrently valid; and (3) it furnishes a structured format conducive to automated evaluation and scoring. Each question was meticulously designed to include six options, distinctly labeled A through F.

\subsubsection{Generation Process}
The generation of questions was autonomously driven by the Gemini Pro Vision API, leveraging the extracted text snippets as contextual input. A detailed prompt, provided to the API, instructed the model to execute the following steps:
\begin{enumerate}
    \item Generate five multiple-choice questions pertinent to the provided text snippet, concentrating on core aerospace manufacturing knowledge deemed essential for experts in the field, even in the absence of the specific text. This emphasized the requirement for fundamental domain expertise.
    \item Furnish six distinct options (A-F) for each question, ensuring diversity and plausibility.
    \item Explicitly indicate one or more correct answers for each question, providing definitive answer keys.
    \item Incorporate a detailed explanation for each correct answer, meticulously justifying the selection based on established aerospace manufacturing principles and domain knowledge.
    \item Categorize each question according to its difficulty level (easy, medium, or hard), adhering to a distribution of one easy, two medium, and two hard questions per snippet. The prompt explicitly directed the inclusion of potential distractors in harder questions, such as partially correct statements or erroneous parameter values, thereby simulating the inherent complexities of real-world decision-making in manufacturing environments.
\end{enumerate}
The generated questions were systematically structured in JSON format to ensure ease of parsing and facilitate subsequent automated evaluation procedures. The question generation process incorporated techniques such as keyword extraction, sentence paraphrasing, and the strategic creation of plausible yet incorrect answer options (distractors) to effectively assess the LLMs' discriminatory capabilities.

\subsubsection{Expert Review and Validation}
Subsequent to the automated generation phase, a sampling of approximately 300 questions was meticulously selected from the generated question pool. This sampled subset was then subjected to rigorous review by a panel of seasoned experts in the field of aerospace manufacturing. This expert review process served a critical validation function, ensuring the generated questions' accuracy, relevance, and appropriateness for evaluating domain expertise. The experts assessed each question for: (1) factual correctness and alignment with established aerospace manufacturing principles; (2) clarity and unambiguousness of question wording and options; (3) relevance to practical aerospace manufacturing knowledge; and (4) appropriate difficulty level. Feedback from the expert review was instrumental in refining question wording, correcting any factual inaccuracies, and ensuring the overall quality and validity of the final question set used for LLM evaluation.

\subsubsection{Difficulty Control}
Controlling the difficulty of the generated questions was primarily achieved through a carefully orchestrated prompting strategy employed with the Gemini Pro Vision API. The prompt explicitly instructed the model to generate questions spanning varying difficulty levels, providing specific guidance on crafting more challenging questions that incorporated nuanced options and potentially misleading distractors. The intended difficulty gradation was predicated upon the depth of knowledge demanded, the intrinsic complexity of the concepts interrogated, and the subtlety of the incorrect options presented, effectively mirroring the cognitive demands encountered by professionals within the aerospace manufacturing domain.

\subsection{Evaluation Metrics}

To quantitatively assess the performance of the LLMs, we employed a comprehensive suite of evaluation metrics. This included a custom scoring system meticulously designed to reflect the paramount importance of accuracy within aerospace manufacturing.

\begin{itemize}
    \item \textbf{Custom Scoring System:} Recognizing the critical implications and potential ramifications of inaccurate information in aerospace manufacturing, we implemented a bespoke scoring mechanism for individual questions.  In this system, selecting a correct option accrues +1 point, while incorrectly selecting an option incurs a penalty of -3 points. Options left unselected contribute neither positively nor negatively to the score. This scoring system is strategically formulated to strongly incentivize the selection of only unequivocally correct answers and to penalize speculative responses or the inclusion of erroneous information. This methodological choice aims to mirror the real-world consequences where inaccurate answers in manufacturing contexts can precipitate significant errors and financial burdens. The aggregate score, and consequently the derived average score, provides a robust metric for evaluating the model's reliability within this critical domain.

    \item \textbf{Overall Accuracy:}  This metric represents the percentage of questions for which the LLM accurately identified all and only the correct options, achieving a perfect match with the pre-defined answer key. It serves as a stringent measure of absolute accuracy.

    \item \textbf{Attempt Rate:}  This is calculated as the ratio of questions for which the LLM provided a response (by selecting at least one option) to the total number of questions. This metric indicates the model's engagement propensity with the posed questions.

    \item \textbf{Average Score:} This metric is derived by dividing the total score attained by the LLM across all questions by the total number of questions. Informed by our custom scoring system, the average score offers a holistic perspective on the model's performance, balancing correct selections against incorrect ones and providing a nuanced understanding of its overall reliability.

    \item \textbf{Category-Specific Metrics:} To gain insights into performance variations across different facets of aerospace manufacturing knowledge, we computed the overall accuracy, attempt rate, and average score for each question category (e.g., Aerospace Assembly, Aerospace Materials).
\end{itemize}

The judicious selection of the custom scoring system underscores the imperative for high fidelity and the need to discourage the selection of incorrect options in a domain where errors can engender significant repercussions. The average score, derived from this system, furnishes a valuable indicator of the model's reliability and suitability for knowledge-intensive tasks in aerospace manufacturing.

\subsection{Experimental Procedure}

The experimental evaluation was executed according to the following procedural steps:

\begin{enumerate}
    \item \textbf{Question Set Preparation}: A comprehensive question set was generated following the methodology delineated in Section 3.2. This generation process yielded approximately 7500 questions, categorized into three principal types based on the source material: Aerospace Assembly, Aerospace Materials (encompassing coatings, plating, and textiles), and Structural Steel. The question count within each category exhibited variance, with Aerospace Assembly constituting the largest question pool.
    \item \textbf{LLM Selection}: A diverse cohort of LLMs was selected for evaluation, including GPT-4o, GLM-4-Air, DeepSeek-Chat-V3, and others. The specific versions of each model were meticulously documented to ensure experimental reproducibility.
    \item \textbf{Batch Processing}: The generated questions were processed in batches of 5 for efficient API utilization. For each batch, a prompt was constructed and submitted to each selected LLM via their respective APIs. This batch processing strategy was implemented to enhance efficiency and effectively manage API rate limitations.
    \item \textbf{Response Recording}: The raw responses from each LLM, including the selected options and any accompanying explanations, were systematically recorded and stored in JSONL format for subsequent in-depth analysis. This process ensured data integrity and facilitated downstream processing.
    \item \textbf{Automated Accuracy Evaluation}: The accuracy of each LLM's responses was automatically evaluated by computationally comparing the selected options against the correct answers specified within the question set. This automated evaluation process also calculated the evaluation metrics described in Section 3.3, inclusive of the custom score. Strict adherence to the predetermined JSON format in the prompts was crucial to enable automated parsing of the LLMs' responses. Error handling mechanisms were incorporated to manage instances where LLM responses deviated from the anticipated format, with attempts made to rectify malformed JSON outputs.
    \item \textbf{Performance Analysis and Reporting}: The collected results were subsequently aggregated and analyzed. This involved computing the overall accuracy, attempt rate, average score, and category-specific metrics for each LLM.  The analysis culminated in the generation of comprehensive reports in both Markdown and Excel formats, providing detailed breakdowns of model performance across all evaluated metrics and categories.
\end{enumerate}

The entirety of the experimental setup, encompassing the specific models employed, API call parameters, and automated analysis procedures, was meticulously documented to guarantee the transparency and replicability of our findings. The adoption of automated evaluation and reporting mechanisms facilitated a rigorous and efficient assessment of LLM performance concerning aerospace manufacturing knowledge.

\subsection{Overall Performance}

Table \ref{tab:overall_performance} presents the comprehensive performance evaluation of the LLMs across the entire question set, encompassing all categories. As the data illustrates, the models exhibited a spectrum of proficiency in addressing the specialized aerospace manufacturing queries.

\begin{table*}[t]
    \centering
    \caption{Overall Performance of Evaluated LLMs}
    \label{tab:overall_performance}
    \begin{tabular}{lccccc}
        \toprule
        \textbf{Model} & \textbf{Total Questions} & \textbf{Attempted Questions} & \textbf{Overall Accuracy} & \textbf{Total Score} & \textbf{Average Score} \\
        \midrule
        gemini-2.0-flash-exp & 2480 & 2461 & 50.75\% & 2904.0 & 1.17 \\
        claude-3-5-sonnet-20241022 & 2480 & 2467 & 49.37\% & 2595.0 & 1.05 \\
        deepseek-chat-v3 & 2480 & 2474 & 45.77\% & 2525.0 & 1.02 \\
        doubao-pro-32k & 2480 & 2466 & 45.66\% & 2512.0 & 1.01 \\
        gpt-4o & 2480 & 2474 & 44.00\% & 2438.0 & 0.98 \\
        Qwen2.5-72B-Instruct & 2464 & 2463 & 43.83\% & 1429.0 & 0.58 \\
        glm-4-plus & 2480 & 2445 & 43.79\% & 1361.0 & 0.55 \\
        ERNIE-4.0-8K & 2480 & 2472 & 42.92\% & 1985.0 & 0.80 \\
        Llama-3.3-70B-Instruct & 2451 & 2443 & 39.53\% & 1660.0 & 0.68 \\
        glm-4-air & 2480 & 2430 & 38.99\% & 1193 & 0.48 \\
        gemini-2.0-flash-thinking-exp-1219 & 2480 & 717 & 12.82\% & 398.0 & 0.16 \\
        \bottomrule
    \end{tabular}
\end{table*}

The data presented in Table \ref{tab:overall_performance} reveals that Gemini-2.0-flash-exp and Claude-3-5-sonnet-20241022 achieved the highest overall accuracy rates, at 50.75\% and 49.37\% respectively.  These models also attained the peak average scores, with Gemini-2.0-flash-exp at 1.17 and Claude-3-5-sonnet-20241022 at 1.05, indicating a more favorable balance between correct and incorrect option selections within our custom scoring framework. GPT-4o exhibited a comparable overall accuracy of 44.00

It is pertinent to note that while all models were presented with the complete question set of 2480 questions, the "Attempted Questions" count varies slightly across models. This variation is primarily attributed to challenges encountered during JSON response parsing and automated data recording. Instances of malformed JSON responses, despite automated repair attempts, occasionally resulted in the inability to accurately log model answers. This aspect, while seemingly a technical artifact of data collection, can also be interpreted as reflecting a facet of model robustness in real-world application scenarios where output format consistency cannot be guaranteed.

It is noteworthy that Gemini-2.0-flash-thinking-exp-1219 exhibits a markedly lower count of "Attempted Questions" in comparison to the other evaluated models. While all models were presented with the complete question set of 2480 questions, Gemini-2.0-flash-thinking-exp-1219's responses included a detailed exposition of its reasoning process alongside the answer itself. This verbose output, incorporating the model's internal "thinking process," posed substantial challenges for our automated evaluation pipeline, which relied on consistent JSON-formatted responses for data recording.  Specifically, the inclusion of this extensive reasoning made reliable parsing and extraction of valid JSON answers exceptionally difficult, leading to significant data recording inconsistencies. Consequently, a considerable number of the model's responses could not be accurately logged as "Attempted Questions." Therefore, the performance metrics presented for Gemini-2.0-flash-thinking-exp-1219 should be interpreted cautiously, acknowledging that they are based on a dataset with fewer recorded responses due to these JSON parsing challenges, rather than reflecting the model's complete performance across the entire question set.

\section{Performance by Question Category}

To gain a more nuanced understanding of the LLMs' capabilities and limitations, we proceeded with a category-specific analysis of their performance across the three defined question categories.

\subsection{Aerospace Assembly}

The Aerospace Assembly category encompassed a broad spectrum of knowledge concerning airplane assembly procedures. Table \ref{tab:AerospaceAssembly_performance} presents the performance metrics for each LLM specifically within the Aerospace Assembly question set.

\begin{table*}[t]
    \centering
    \caption{Performance on Aerospace Assembly Questions}
    \label{tab:AerospaceAssembly_performance}
    \begin{tabular}{lccccc}
        \toprule
        \textbf{Model} & \textbf{Total Questions} & \textbf{Attempted Questions} & \textbf{Accuracy} & \textbf{Total Score} & \textbf{Average Score} \\
        \midrule
        gemini-2.0-flash-exp & 1600 & 1599 & 52.66\% & 2009.0 & 1.26 \\
        claude-3-5-sonnet-20241022 & 1600 & 1599 & 52.35\% & 1804.0 & 1.13 \\
        doubao-pro-32k & 1600 & 1591 & 49.87\% & 1841.0 & 1.16 \\
        deepseek-chat-v3 & 1600 & 1594 & 49.31\% & 1846.0 & 1.16 \\
        gpt-4o & 1600 & 1594 & 47.80\% & 1825.0 & 1.14 \\
        Qwen2.5-72B-Instruct & 1600 & 1597 & 47.21\% & 1273.0 & 0.80 \\
        glm-4-plus & 1600 & 1569 & 46.02\% & 1099.0 & 0.70 \\
        ERNIE-4.0-8K & 1600 & 1593 & 45.86\% & 1509.0 & 0.95 \\
        Llama-3.3-70B-Instruct & 1600 & 1595 & 42.56\% & 1366.0 & 0.86 \\
        glm-4-air & 1600 & 1568 & 41.79\% & 1065 & 0.67 \\
        gemini-2.0-flash-thinking-exp-1219 & 1600 & 222 & 13.88\% & 53.0 & 0.24 \\
        \bottomrule
    \end{tabular}
\end{table*}

Consistent with the overall performance trends, Gemini-2.0-flash-exp and Claude-3-5-sonnet-20241022 again exhibited the highest accuracy within the Aerospace Assembly category, achieving 52.66\% and 52.35

\subsection{Aerospace Materials}

The Aerospace Materials category featured questions focused on a diverse array of materials pertinent to aerospace engineering, including titanium alloys, coatings, fuels, and lubricants. The performance outcomes for this category are detailed in Table \ref{tab:AerospaceMaterials_performance}.

\begin{table*}[t]
    \centering
    \caption{Performance on Aerospace Materials Questions}
    \label{tab:AerospaceMaterials_performance}
    \begin{tabular}{lccccc}
        \toprule
        \textbf{Model} & \textbf{Total Questions} & \textbf{Attempted Questions} & \textbf{Accuracy} & \textbf{Total Score} & \textbf{Average Score} \\
        \midrule
        gemini-2.0-flash-exp & 500 & 483 & 42.92\% & 376.0 & 0.75 \\
        claude-3-5-sonnet-20241022 & 500 & 488 & 40.98\% & 391.0 & 0.78 \\
        deepseek-chat-v3 & 500 & 500 & 35.40\% & 227.0 & 0.45 \\
        doubao-pro-32k & 500 & 498 & 32.33\% & 201.0 & 0.40 \\
        gpt-4o & 500 & 500 & 32.00\% & 169.0 & 0.34 \\
        ERNIE-4.0-8K & 500 & 500 & 33.00\% & 117.0 & 0.23 \\
        Qwen2.5-72B-Instruct & 500 & 497 & 31.93\% & -120.0 & -0.24 \\
        Llama-3.3-70B-Instruct & 500 & 490 & 30.61\% & 23.0 & 0.05 \\
        glm-4-plus & 500 & 496 & 35.89\% & -29.0 & -0.06 \\
        glm-4-air & 500 & 490 & 29.33\% & -158 & -0.32 \\
        gemini-2.0-flash-thinking-exp-1219 & 500 & 348 & 27.68\% & 206.0 & 0.41 \\
        \bottomrule
    \end{tabular}
\end{table*}

In concordance with the aggregate performance analysis, performance across all models exhibited a discernible decline within the Aerospace Materials category when contrasted with Aerospace Assembly. DeepSeek-Chat-V3 achieved the highest accuracy at 35.40

\subsection{Structural Steel}

The Structural Steel category specifically targeted knowledge pertaining to commonly utilized structural steel alloys in aerospace engineering. Table \ref{tab:StructuralSteel_performance} provides a detailed exposition of the performance outcomes for each LLM within this category.

\begin{table*}[t]
    \centering
    \caption{Performance on Structural Steel Questions}
    \label{tab:StructuralSteel_performance}
    \begin{tabular}{lccccc}
        \toprule
        \textbf{Model} & \textbf{Total Questions} & \textbf{Attempted Questions} & \textbf{Accuracy} & \textbf{Total Score} & \textbf{Average Score} \\
        \midrule
        gemini-2.0-flash-exp & 380 & 379 & 52.89\% & 519.0 & 1.37 \\
        claude-3-5-sonnet-20241022 & 380 & 380 & 47.89\% & 400.0 & 1.05 \\
        doubao-pro-32k & 380 & 377 & 45.53\% & 470.0 & 1.25 \\
        deepseek-chat-v3 & 380 & 380 & 44.74\% & 452.0 & 1.19 \\
        gpt-4o & 380 & 380 & 43.95\% & 444.0 & 1.17 \\
        glm-4-plus & 380 & 380 & 45.26\% & 291.0 & 0.77 \\
        ERNIE-4.0-8K & 380 & 379 & 43.68\% & 359.0 & 0.94 \\
        glm-4-air & 380 & 372 & 42.74\% & 287.0 & 0.76 \\
        Llama-3.3-70B-Instruct & 380 & 375 & 38.32\% & 271.0 & 0.74 \\
        Qwen2.5-72B-Instruct & 380 & 379 & 45.26\% & 276.0 & 0.75 \\
        gemini-2.0-flash-thinking-exp-1219 & 380 & 147 & 38.68\% & 139.0 & 0.37 \\
        \bottomrule
    \end{tabular}
\end{table*}

In the Structural Steel category, Gemini-2.0-flash-exp again reached the highest average score, at 1.37, closely trailed by Claude-3-5-sonnet-20241022 (1.05) and GPT-4o (1.17). The accuracy rates within this category are broadly consistent with the overall accuracy and the Aerospace Assembly category, indicating that the models possess a moderate level of proficiency in structural steel concepts. The positive average scores across all models in this category suggest an improved ability to differentiate correct answers when compared to the more challenging Aerospace Materials domain.

The synthesized analysis across these distinct categories reinforces the observation that while a majority of models demonstrate a reasonable level of comprehension regarding Aerospace Assembly and Structural Steel knowledge, the Aerospace Materials domain presents a more pronounced challenge. The average score, derived through our custom penalty-based system, continues to serve as a discriminating metric for assessing model reliability, effectively distinguishing those models that are not only frequently accurate but also demonstrably more averse to selecting incorrect options—a critical consideration for deployment in high-stakes sectors like aerospace manufacturing.

\section{Discussion}

The experimental results offer valuable insights into the current capabilities and limitations of LLMs when applied to the specialized domain of aerospace manufacturing. The high attempt rates across all models indicate a willingness to engage with the domain-specific questions. However, the varying levels of overall accuracy and average scores reveal significant differences in their comprehension and reliability within this critical field.

Notably, Gemini-2.0-flash-exp and Claude-3-5-sonnet-20241022 consistently demonstrated the highest overall accuracy and average scores, particularly in the \textit{zhuangpei} (airplane assembly) category. This suggests a potentially stronger understanding of assembly processes compared to other models. The comparatively lower performance in the \textit{cailiao} (aerospace materials) category across all models highlights the inherent complexity and nuanced knowledge required in material science. The negative average scores observed for some models in this category further underscore the risk of inaccurate information generation when dealing with intricate material properties and applications.

The custom scoring system, with its penalty for incorrect selections, proved effective in differentiating models based on their reliability. The average score provides a more nuanced measure than simple accuracy, reflecting not only the frequency of correct answers but also the caution exercised in avoiding incorrect ones. This is particularly crucial in aerospace manufacturing, where the cost of misinformation can be substantial.

Comparing the performance across different question categories, it is evident that LLMs exhibit varying levels of proficiency in different sub-domains of aerospace manufacturing. While assembly processes and structural steel concepts appear to be relatively better understood, the intricacies of aerospace materials present a significant challenge. This suggests that the training data for these models may have a bias towards certain aspects of manufacturing or lacks sufficient depth in the specialized knowledge of material science within the aerospace context.

The findings align with existing literature highlighting the issue of "hallucination" in LLMs, especially in specialized domains. While LLMs possess impressive natural language processing abilities, their factual knowledge and reasoning capabilities in niche areas like aerospace manufacturing still require significant improvement. This study emphasizes the importance of rigorous, domain-specific evaluation to assess the suitability of LLMs for safety-critical applications.

The potential applications of LLMs in aerospace manufacturing, such as assisting in process design or information retrieval, are promising. However, our results caution against relying solely on these models without careful human oversight and verification, especially in areas requiring precise technical knowledge like material selection.

\section{Conclusion and Future Work}

This paper presented a comprehensive evaluation of several prominent Large Language Models on a specialized question bank derived from authoritative aerospace manufacturing resources. Our methodology, incorporating automated question generation and a custom scoring system, provides a valuable framework for assessing the accuracy and reliability of LLMs in this demanding domain.

The experimental results reveal a spectrum of performance across the evaluated models, with some demonstrating a stronger grasp of certain aerospace manufacturing concepts than others. The challenges observed in the aerospace materials category underscore the need for further advancements in LLMs' ability to handle complex, domain-specific knowledge. The custom scoring system effectively highlighted the models that exhibit a more cautious and reliable approach to answering questions in this safety-critical field.

Our findings emphasize that while LLMs hold potential for various applications in aerospace manufacturing, their current limitations, particularly the propensity for generating inaccuracies, necessitate careful consideration and rigorous validation before deployment in critical tasks.

Future work will focus on several key areas:
\begin{itemize}
    \item \textbf{Expanding the Evaluation Dataset:} We aim to broaden the scope of our evaluation dataset by incorporating more diverse question types and covering a wider range of aerospace manufacturing sub-domains. This includes integrating questions that assess problem-solving and decision-making skills in practical manufacturing scenarios.
    \item \textbf{Investigating Retrieval-Augmented Generation (RAG):} Exploring the effectiveness of RAG techniques to improve the accuracy of LLM responses by grounding them in relevant aerospace manufacturing documentation and knowledge bases. This approach could mitigate the issue of hallucinations and enhance the reliability of generated information.
    \item \textbf{Fine-tuning LLMs on Aerospace Manufacturing Data:} Investigating the impact of fine-tuning pre-trained LLMs on curated aerospace manufacturing datasets to enhance their domain-specific knowledge and reasoning capabilities.
    \item \textbf{Developing More Granular Evaluation Metrics:} Exploring more fine-grained evaluation metrics to assess specific aspects of LLM performance, such as their ability to understand technical drawings, interpret industry standards, and provide justifications for their answers based on established principles.
    \item \textbf{Evaluating the Impact of Question Difficulty:} Conducting a more detailed analysis of model performance across different difficulty levels to identify specific areas where models struggle and to refine the question generation process.
\end{itemize}

Ultimately, our goal is to contribute to the responsible and effective integration of LLMs into the aerospace manufacturing industry, ensuring that these powerful tools are deployed in a manner that enhances efficiency and safety without compromising the stringent quality standards of the sector.

\section*{Acknowledgment}

The authors gratefully acknowledge the invaluable contributions of intern Siteng Hu, whose exceptional dedication and expertise were instrumental in the success of this research. Siteng Hu played a pivotal role in establishing the experimental framework, conducting comprehensive technical processing tasks, and performing in-depth data analysis that significantly enhanced the robustness of our findings. Furthermore, his insightful feedback and meticulous efforts in revising and refining the manuscript were crucial in shaping the clarity and coherence of this work. His multifaceted contributions have left a lasting impact on both the experimental and analytical dimensions of this study.



%

\end{document}